# Semantic Similarity Measure of Natural Language Text through Machine Learning and a Keyword-Aware Cross-Encoder-Ranking Summarizer - A Case Study Using UCGIS GIS&T Body of Knowledge

Yuanyuan Tian[1], Wenwen Li[1,*], Sizhe Wang[1], Zhining Gu[1]

[1]School of Geographical Sciences and Urban Planning, Arizona State University, Arizona, Tempe, USA
[*]Corresponding Author: wenwen@asu.edu

**Abstract**: Initiated by the University Consortium of Geographic Information Science (UCGIS), GIS&T Body of Knowledge (BoK) is a community-driven endeavor to define, develop, and document geospatial topics related to geographic information science and technologies (GIS&T). In recent years, GIS&T BoK has undergone rigorous development in terms of its topic re-organization and content updating, resulting in a new digital version of the project. While the BoK topics provide useful materials for researchers and students to learn about GIS, the semantic relationships among the topics, such as semantic similarity, should also be identified so that a better and automated topic navigation can be achieved. Currently, the related topics are either defined manually by editors or authors, which may result in an incomplete assessment of topic relationship. To address this challenge, our research evaluates the effectiveness of multiple natural language processing (NLP) techniques in extracting semantics from text, including both deep neural networks and traditional machine learning approaches. Besides, a novel text summarization - KACERS (Keyword-Aware Cross-Encoder-Ranking Summarizer) - is proposed to generate a semantic summary of scientific publications. By identifying the semantic linkages among key topics, this work provides guidance for future development and content organization of the GIS&T BoK project. It also offers a new perspective on the use of machine learning techniques for analyzing scientific publications, and demonstrate the potential of KACERS summarizer in semantic understanding of long text documents.

## 1. INTRODUCTION

A Body of Knowledge (BoK) defines important knowledge topics to complete a job or task in a specific domain, and contributes to professional development needs (Stelmaszczuk-Gorska et al., 2020). Expected users of a BoK include educators, students, and professionals. GIS related BoK works can assist GIS domain curriculum planning and revision, program assessment and articulation, professional certification, and employee screening (Prager & Plewe, 2009; C. Wang et al., 2020). For example, GIS&T Body of Knowledge, or GIS&T BoK for short, represents the Geographic Information Science and Technology domains in a hierarchical fashion of knowledge areas, units, and topics. GIS&T BoK project is adopted by the American Association of Geographers as a standard GIScience learning guideline (DiBiase et al., 2007). Encyclopedia-style BoKs have begun to include explicit cross-references or a list of related topics at the end of each topic, which makes it easier for knowledge seekers to explore related learning materials. This practice is common, just like in Wikipedia's "see also" section, the "cross-reference" section also exists in the Encyclopedia of GIS (Shekhar & Xiong, 2007), Encyclopedia of Geographic Information Science (K. Kemp, 2008), International Encyclopedia of Geography (Brunn, 2019) , and GIS&T BoK.

Topic relevance can be manually crafted, but it is possible to automate the process by Natural Language Processing (NLP). NLP is a technique to process and analyze language data, aiming to extract meanings (semantics) from text (Nadkarni et al., 2011). In recent years, the advances in Geospatial Artificial Intelligence (GeoAI) (Li, 2020, 2022) have further empowered NLP to utilize advanced AI technology especially deep learning to process and understand natural language text. For instance, one of the main research applications of NLP is to measure the semantic similarity among scientific publications for tasks such as cross-referencing and topic recommendation (Gargiulo et al., 2018; Nagwani, 2015). In recent years, the development of large language models and other deep learning models pretrained on the large corpus have brought significant advances for NLP. These new models are capable of learning universal language representations, which are beneficial for a variety of downstream NLP tasks such as semantic similarity measurement (Qiu et al., 2020). Semantic similarity measures the degree to which two textual pieces carry similar meaning and this measure has been widely used in the GIS domain to help find related geographic terms (Ballatore et al., 2013), match spatial entity descriptions (Ma et al., 2018), identify similar spatial features (Li et al., 2012), leverage spatial context semantics (B. Wang et al., 2020), and so on.

However, the semantic similarity of GIS topics, knowledge, and publications has not been extensively studied and many of such tasks remain relying on manual work (Frazier et al., 2018). On the one hand, as it always involves large amount of time for someone to read publications with long texts and understand its semantics, the task for manual determination of related topics is therefore effort-intensive and may not be complete. For example, when contributing to GIS&T BoK, the related topic list is provided by a topic author who may or may not have a comprehensive knowledge of all the topics in the project. Thus, there is a desire to improve the process of identifying relevant topics for cross-referencing in the BoKs, especially with the support and automation from NLP. A few works started to investigate semantics of GIS related BoKs, for example, linking those BoKs with scientific papers (Du et al., 2021). However, understanding semantics within a BoK itself is also worth studying but yet to be exploited, and find appropriate NLP models to deal with lengthy BoK contents is challenging.

The goal of this work is to evaluate the effectiveness of different NLP techniques in automated extracting semantics from GIS BoK texts and uncover topic relevance. It is important to note that although GIS&T BoK is used as the case study in our research, the methodology is generalizable and can be applied to analyze other scientific publications to extract core semantic representations of each article. In our research, we analyze the commonalities and differences between machine-calculated relevance and human rankings. By comparing multiple NLP techniques, including traditional approaches and deep neural network models, our work can provide both the authors and audience of the GIS&T BoK with more accurate content recommendation. Besides, a novel summarization model - Keyword-Aware Cross-Encoder-Ranking Summarizer (KACERS) is proposed to extract semantics from long texts effectively. For encyclopedia style knowledge materials using very closely related domain language (e.g., GIS domain), our approach is useful in obtaining a good coverage of semantic relevance without the need to manually edit the topics. This work also shows the potential of advanced deep learning models to gain new insights on semantic analytics of scientific publications.

The remainder of the article is organized as follows. Section 2 reviews related works in the literature. Section 3 presents the data, methods, and evaluation metrics used in this study. Section 4 presents and analyzes the results. Section 5 concludes this work and discuss future research directions.

# 2. LITERATURE REVIEW

In this section, we first review studies on GIS related BoKs. Next, we review literatures on natural language processing and semantic similarity measures. Then, text summarization as a key technique for generating semantically condensed text is reviewed.

## 2.1 GIS related BoK research

There are many ongoing or completed GIS related BoK works. The first version of GIS&T BoK was introduced in 2006 by UCGIS with contributions from GIS researchers and with support of Esri to establish what at the time current and aspiring geospatial professionals needed to know and be able to implement in terms of geospatial theories, methods and tools (DiBiase et al., 2007). Initiated in 2013 and discussed during 2014 UCGIS Symposium, several activities yielded a new project to initiate the second edition of GIS&T Bok in a web-based-environment fashion (Waters, 2013; Wilson, 2014). The latest GIS&T BoK aims to create a more comprehensive reference work which covers both the classic theory and the new technology for GIS students and practitioners.

There are several BoK efforts similar to GIS&T BoK. The United States Geospatial Intelligence Foundation conducts an analysis on cross-industry jobs, and produces Geospatial Intelligence (GEOINT) Essential Body of Knowledge (EBK) to identify the knowledge, skills, and abilities critical to the GEOINT workforce (DiBiase et al., 2007). In Europe, the Earth Observation and Geographic Information (EO4GEO) BoK is developed (Hofer et al., 2020; Stelmaszczuk-Gorska et al., 2020). The EO4GEO is re-engineered and expanded on the structure of GIS&T BoK and is focused on skill building to fill the gap between the supply and demand of education and training in space and geospatial sciences (Ahearn et al., 2013). In South Africa, a university course framework (du Plessis & Van Niekerk, 2014) was developed and customized to meet local and international requirements based on GIT&T BoK. These projects reflect great interests in GIS related BoKs development by various countries, organizations, and experts.

Previous works on GIS BoK mainly focus on curriculum design (Blanford et al., 2021; M. DeMers, 2019; Elwood & Wilson, 2017; Hossain & Reinhardt, 2012; A. B. Johnson & Sullivan, 2010; K. K. Kemp, 2012; Prager, 2012) and skill identification (Hong, 2016; A. Johnson, 2019; Sadvari, 2019; Wallentin et al., 2015). Only a few semantic analysis works have been conducted to understand geospatial domain knowledge and essential terminologies. Actionable verbs from the learning objectives of each GIS&T BoK topic has semantic values in evaluating cognitive levels of each BoK entries (M. N. DeMers, 2009). One advantage of GIS&T BoK is its depth and breadth since it covers a few hundred topics core to GIScience research and education. However, this also brings challenges when finding patterns and semantic associations between topics in practice. A structured visualization of GIS&T BoK as colored cells in row for curriculum assessment and evaluation (Prager & Plewe, 2009) is helpful to identify a revision need, but an enhancement could be made possible by revealing semantically related entities, so we can spot entities reflecting possible reasons of poor teaching evaluations or potential impacts of curriculum changes. This can be achieved through semantic similarity analysis. Content analysis is a popular method applied on GIS&T BoK to support curriculum design. Frazier et al. (2018) developed a method to match headings and subheadings of GIS textbooks with GIS&T BoK topics. However, semantics of those contents were captured purely manually, which is to some degree subjective and labor-intensive. Such works can potentially benefit from automated semantic similarity analysis.

## 2.2 NLP and semantic similarity

Modeling semantic similarity between textual components is an important NLP research problem, and has gained significant attentions by researchers from a variety of scientific domains, such as biology, geology, hydrology, and urban planning (Pesquita et al., 2009; Zhang et al., 2021; C. Zhuang et al., 2021; D. Zhuang et al., 2020). The goal is to determine to what degree two textual components are semantically similar in terms of their content. Semantic similarity analysis extracts the semantic relationships among a collection of words, sentences, paragraphs, or documents. The advantage of employing machine learning in this process is to enable analytical capabilities that go beyond an individual person's understanding of texts. Semantic similarity is useful in various NLP tasks such as information retrieval, machine translation, question answering, entity resolution and beyond (Ebraheem et al., 2018; Li et al., 2019; Varelas et al., 2005; Yih et al., 2014; Zou et al., 2013). In practice, it has a wide range of real life applications such as patent class prediction, scientific text comparison, newspaper article similarity analysis, topic discovery for United Nations speeches, and concept derivation for encyclopedias (Geum & Kim, 2020; Gomez, 2019; Gong et al., 2019; Shalaby & Zadrozny, 2017; Watanabe & Zhou, 2020). Going beyond determining the semantic similarity as a binary decision (1 or 0, as similar or not), generating non-binary scores of text pairs followed by ranking has become a common goal of semantic similarity analysis. For example, a numeric score is expected to compare with human labeled score of 0-5 for sentence representation of popular benchmarks such as the semantic textual similarity (STS) tasks from 2012 to 2016 (Conneau & Kiela, 2018). Although there are different measures of similarity, the cosine similarity has been a widely adopted approach (Mohammad & Hirst, 2012).

Categorized by surveys on semantic similarity techniques (Chandrasekaran & Mago, 2021; Elavarasi et al., 2014), there are three major classes including knowledge-based methods, corpus-based methods, and deep neural network-based methods. The knowledge-based methods are based on predefined knowledge sources such as ontologies, dictionaries or thesauri and ontological relationships between words. Those methods require underlying sources to be well structured and formatted. On the other hand, corpus-based methods utilize information retrieved from large corpora. These approaches are welcomed and dominate recent works as they are free from being highly dependent on predefined knowledge sources. This class stems from the distributional hypothesis assuming that similar words frequently appear together (McDonald & Ramscar, 2001). A very important notion, word embeddings, is developed as a powerful approach in this class. Word embeddings provides semantically meaningful representations of words by vectors (Kusner et al., 2015). There are various ways to compute word embeddings, such as shallow neural networks (Mikolov et al., 2013) and word co-occurrence matrix (Dumais, 2004; Pennington et al., 2014). Recently, with the increasing availability of large datasets and computing power, deep learning has actively pushed the boundary of semantic similarity research, and forms the third class as mentioned above. A neural network architecture known as Transformer (Vaswani et al., 2017) adopts the mechanism of self-attention to increase model awareness of context in sequential data such as texts. An example is Bidirectional Encoder Representations from Transformers (BERT) developed by Google (Devlin et al., 2018). Its key technical innovation is to leverage the bidirectional training of transformers to learn semantic dependencies of text elements (words, phrases, sentences and paragraphs) from both forward and backward directions. The masked language model and next structure prediction along with a huge amount of training data and Google's immerse computing power distinguish BERT as a milestone development of sequence-to-sequence learning models (Yates et al., 2021). Transformer architectures, such as BERT, are therefore becoming the cutting-edge for deep semantic understanding of text documents of various length.

## 2.3 Text summarization

Comparing semantic similarities between texts is a non-trivial task, especially when texts are long. Text summarization is a useful technique because it can distill main ideas of the lengthy input into a shorter text (Maybury, 1999). There are two major approaches of text summarization: extractive and abstractive approaches (El-Kassas et al., 2021). Extractive approaches select the most informative sentences from the original text, while abstractive approaches paraphrase the main idea by generating new sentences. It is easy for some classic extractive approaches such as LexRank (Erkan & Radev, 2004) to utilize word embeddings for sentence scoring then identify central sentences. There are advanced neural network models applied on the abstractive approach such as sequence-to-sequence models which consider the input text as a sequence of tokens, and the output summary is a shorter sequence of tokens (Lewis et al., 2019; Shi et al., 2021). Extractive approach leads to a higher accuracy because selected sentences are directly extracted from the text documents so the terms have exact appearance in the original text. Abstractive approach generates a more flexible summary, but the quality still needs to be improved because it involves generation of new text, and this remains a developing field. Several efforts have been made on scientific publication summarization with either approach or even hybrid approaches combining both strategies (Cagliero & La Quatra, 2020; Erera et al., 2019; Subramanian et al., 2019). However, very few works have been done in the GIScience domains (Gregory et al., 2015).

To address this issue, our work develops a novel text analysis framework that can integrate multiple NLP methods for the semantic analysis and similarity measure of scientific publications particularly the GIS&T BoK. Next section describes our methodology in detail.

## 3. METHODS

As stated, the goal of this work is to evaluate the effectiveness of different NLP techniques in extracting semantics from GIS knowledge texts to uncover topic relevance. To achieve this, a semantic data processing workflow for scientific publications is developed (Section 3.1). We then introduce a new method to improve text summarization and semantic extraction from long text (Section 3.2). Next, we will introduce different NLP models that can be leveraged for semantic similarity calculation (Section 3.3). Finally, we describe the evaluation metrics used to assess model performance (Section 3.4).

## 3.1 Methodological workflow

BoK entities are complex: (1) they are documents containing several thousand words; (2) they are semi-structured documents with predefined segments, and (3) the language may vary a lot across segments within one entity. Figure 1 illustrates our proposed methodological framework and workflow for processing natural language text especially scientific publications (e.g., BoK entries) for semantic similarity measurement. There are essentially three steps for processing multiple types of scientific publications. The first step is to identify and retrieve contents automatically from the online data source, such as the GIS&T BoK project page. The second step is text preprocessing, semantic modeling and similarity analysis. The third step is result evaluation and analysis.

The starting step (Step 1 in Figure 1) is to understand the structure of publications, then prepare the data ready for analysis. Usually, a topic consists of several segments following an instructed structure, such as abstract, introduction, methodology and conclusions. Segments of interest should be identified for later use. At the end of each topic lays the segment which lists cross-referenced topics. Those related topics are used as ground-truth labels to evaluate the relevance of similar topics identified by the NLP models.

Once the input text is crawled and parsed, it is sent to the core part of the pipeline (Step 2 in Figure 1) for text processing and modeling. Before feeding into the NLP models, the text needs to be preprocessed. Noisy entities and stopping words are removed to clean the text. If there is long text of interest, summarizer can help reduce the length while keeping salient semantics. Next, cleaned text is fed into the NLP models as input. Various NLP models can be employed here, including traditional approaches and deep neural networks, described in Section 3.3. The output is text embeddings, so that the similarity scores of any document pair can be calculated based on the embedded vector to identify semantic relevance.

The results of the semantic analysis can then be evaluated (Step 3 in Figure 1). Some well-acknowledged metrics (Section 3.4) are used to evaluate the performance of difference models using difference text from different segments as inputs. This treatment allows for the calculation of document similarity based on different sections of the publication, helping us gain a better understanding in each section's role and its semantic expressiveness of a topic.

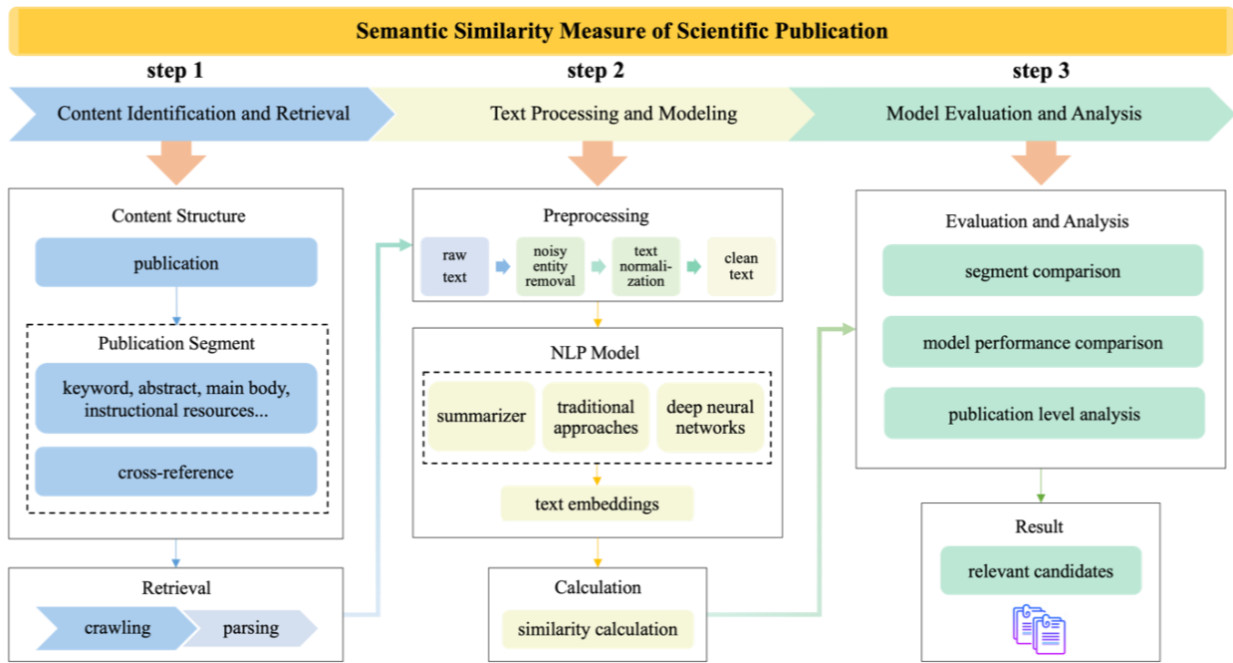

Figure 1. The pipeline to measure the semantic similarity of a body of knowledge.

## 3.2 KACERS: a keyword-aware cross-encoder-ranking summarizer

Because some NLP models required limited length of the input text, for instance, BERT, they are less effective in dealing with the semantic analysis of long text documents. To utilize the power of NLP and at the same time present the most semantically expressive content, a summarizer is applied to find the sentences that represent the most important semantics though out the text. Common objectives of text summarizers are summary generation or further keyword extraction. It has been proved to be effective that adding supporting information such as human comments or aggregated topics could improve quality of the summary (Gao et al., 2019; W. Wang et al., 2019). However, comments could also introduce noises unrelated to the original contents, topics automatically generated by topic models are sets of frequent terms yet so-called topics are not clearly described. By comparison, keywords in the publication are easy to identify and understand by humans, making them a useful supporting information. It is therefore intuitive to take advantage of keywords if available to generate a keyword-related summary, in which sentences related to the defined keywords in the publications will be retrieved. Simple word-matching could locate relevant sentences containing a keyword, while more advanced searching methods based on deep learning, such as cross-encoder (Reimers & Gurevych, 2019), could potentially locate more semantically related sentences. Cross-encoder has the power to identify related sentences by keywords, phrases or questions with a relevance score, which is convenient for ranking. Thus, cross-encoder has potential to fit this semantic summarization task as an engine to search keyword-related sentences and then concatenate selected sentences together as a desired summary.

To achieve good summarization performance, here we propose a novel method to summarize long text based on keywords, named as KACERS (Keyword-Aware Cross-Encoder-Ranking Summarizer). Its workflow is shown in Figure 2. KACER requires two kinds of inputs - a set of keywords, and a piece of long text. First, text is split into sentences. Then keywords and sentences are passed into a transformer-based cross-encoder. For each keyword-sentence pair, two vectors are generated by a cross-encoder (Reimers & Gurevych 2019) to represent a keyword and a sentence, respectively. Different from the vectorization process of a bi-encoder, a cross-encoder simultaneously pays attention to both the keyword and the sentence, whereas a bi-encoder calculate the vector based on limited attention to only one of them. Therefore, the cross-encoder increases the awareness of the keyword in the vector representation of the sentences, and vice versa. For each keyword-sentence pair, a cosine similarity score is calculated to derive how similar a sentence is to a given keyword. Ranking all the possible pairs based on the similarity score, each keyword has found a list of most similar sentences (i.e., top sentences). The t in KACERS-t controls the top t most similar sentences selected for each keyword. Finally, KACERS-t generates a summary of the input text by concatenating these top t sentences found for each given keyword. This new summary contains semantically rich information to represent the original text and is much shorter in length. It is therefore suitable for use in multiple NLP models which requires the input text to be within a certain length.

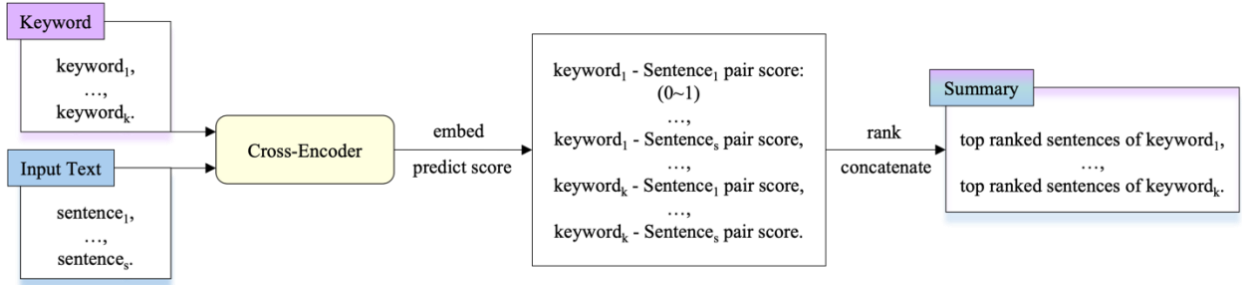

Figure 2. KACERS: Keyword-Aware Cross-Encoder-Ranking Summarizer.

## 3.3 NLP models for semantic similarity

This paper evaluates six popular models to explore the semantic similarity of knowledge topics in Step 2 of our proposed data processing pipeline (Figure 1). We distinguish between two model categories: (1) traditional text-based and shallow machine learning models, and (2) deep learning models. The models derive a semantic representation (i.e., a high-dimensional vector) of different text documents and then similarity measure is performed on these represented vectors.

### 3.3.1 Traditional text-based models and shallow machine learning models

**Latent Semantic Analysis (LSA)**: In natural language processing, TF-IDF (Ramos, 2003) is a fundamental term-weighting model based on word numeric statistics. Instead of pure lexical match like TF-IDF, LSA (Dumais, 2004) improves it by detecting semantic representations from a large corpus of text with the idea of dimensionality reduction (Li et al., 2014). It has some assumptions: meanings of sentences or documentation are given by the aggregation of all words in it; in a large corpus, associations between words can only exist implicitly. For LSA, a term-document matrix is first generated, where each row represents the document, and each column is given by the terms in all the documents. A low-rank approximation technique called truncated singular value decomposition is used to reduce unimportant dimensions of the term-document matrix to make the semantic association better shown in the reproduced matrix.

**Global Vectors (GloVe)**: GloVe (Pennington et al., 2014) is another shallow machine learning model based on word count. It conducts statistics based on both local context and the entire corpus. It utilizes the co-occurrence matrix to make predictions of the probabilities of words in a text to derive its semantic vectors. In another word, it tries to leverage the fact that a certain word is more inclined to be present with another word than any other words in the corpus. GloVe produces dense embedding representations. In terms of computation, due to the addition of the weighting function, it scales well from small corpuses to large ones. Additionally, it caps the occurrence of frequent words (such as the and they) so that the distribution will not be too large and can successfully capture the gist of less frequent but important words.

**Doc2vec**: The idea of Doc2vec in learning paragraph vectors stems from Word2vec (Mikolov et al., 2013), the task of which is to predict certain words given some context words. In Word2vec, context words act as feature vectors during prediction to represent semantics of a specific word. Doc2vec extends this idea toward processing long documents to generate a feature vector to represent each document in the corpus, but it remains a shallow neural network model with a single hidden layer. A popular Doc2vec technique is the model with distributed memory and distributed bag of words (Le & Mikolov, 2014), which is used in our study.

### 3.3.2 Deep learning models

**RoBERTa**: Introduced by Facebook, RoBERTa is a derivative of BERT with improved training methodology such as dynamic masking and more training data than BERT. RoBERTa aims to predict intentionally masked words within unannotated datasets. It mainly involves the modification of hyperparameters in the BERT model by employing dynamic masking and removing the loss function for the next-sentence prediction. Additionally, it trains on larger batches and chooses a different token compression method, i.e., larger byte-pair encoding. As a result, RoBERTa outperforms BERT on several benchmarks such as the General Language Understanding Evaluation (GLUE) (Liu et al., 2019).

**SciBERT** and Scientific Paper Embeddings Citation informed TransformERs (**SPECTER**): SciBERT (Beltagy et al., 2019) and SPECTER (Cohan et al., 2020) apply pretrained BERT models to academic research, using scientific publications as the corpus to further refine the model. SciBERT mainly focuses on the massive corpus of academic publications from Semantic Scholar based on BERT. Full text of papers involving widespread domains is used for model training. It substantially improves the performance of model prediction results from a scientific data - NLP perspective. SPECTER achieves document embeddings of academic publications with the assistance of a transformer language model. It and combines additional scientific domain specific context with the Transformer language models to learn a representation of the original text documents.

Although BERT is a great milestone in NLP, previous research also finds that the architecture of BERT makes it unsuitable for semantic similarity search and clustering (Reimers & Gurevych, 2019). Thus, some efforts have been done to tackle this shortage and improve the performance of BERT related models, and the siamese network architecture they employed helps to generate "semantically meaningful" sentence embeddings. The siamese structure (Schroff et al., 2015) is an artificial neural network that uses the same weights while working in tandem on two different input vectors to compute comparable output vectors. For fine-tuning BERT, siamese network was used to update the weights to make the embeddings semantically meaningful to enable comparison using cosine similarity. Furthermore, SBERT significantly outperformed state-of-the-art sentence embedding methods such as InferSent (Conneau et al., 2017) and Universal Sentence Encoder (Cer et al., 2018) on seven semantic textual similarity tasks. As a family of several BERT related models, RoBERTa and SPECTER in SBERT version are available off the shelf.

### 3.4 Evaluation and analysis metrics

As reviewed, cosine similarity is used to measure the semantic distance between two textual pieces. After text vectors are produced by an NLP model, the cosine of the angle between any two vectors (A and B in Eq. [1]) can be calculated (Landauer et al., 1998), as shown in Equation (1). The result has a value range of [0,1], with 0 indicating the least similar documents and 1 indicating the most similar documents. In the context of BoK semantic modeling, consine similarity is used to measure the similarity between two topics. Each topic is represented as a high-dimensional vector, with each dimension corresponding to a particular semantic feature. Cosine similarity then provides a measure of the cosine of the angle between two vectors, which can be used to determine the similarity between two topics. A cosine similarity score of 1 indicates that two topics are very similar or even identical, while a score of 0 indicates that the two topics are completely dissimilar. In other words, the higher the cosine similarity score, the more similar the two topics are in terms of their semantic features.

$$\cos(\theta) = \frac{A \cdot B}{\|A\|\|B\|} = \frac{\sum_{i=1}^{n} A_i B_i}{\sqrt{\sum_{i=1}^{n} A_i^2} \sqrt{\sum_{i=1}^{n} B_i^2}} \tag{1}$$

As in pipeline Step 3 (Figure 1), to evaluate related topics proposed by those models based on semantic similarities, common metrics of recall, precision, F-measure score, and balanced accuracy are used. These metrics focus on comparisons of the candidate topic set promoted by the model to related topics determined by manual judgements. Although a comprehensive list of pairwise similarities is accessible, many information retrieval studies have demonstrated that users of retrieval systems tend to pay attention mostly to top-ranked results which represent high relevance (Hofmann et al., 2014; Lagun et al., 2014). Here, we largely follow the annotation of NLP common practice (Mitra & Craswell, 2018). For each query topic, top *k* candidate topics ranked by similarity scores, are truncated to use and referred to as *R*. In this sense, $R_q@1$ means the most similar topic, $R_q@10$ means top ten similar topics to a query topic *q*. Recall and precision (Eq. [2] and Eq. [3]) both compute the fraction of relevant topics promoted for *q*, but with respect to the total number of related documents identified by humans as a collection *D*, and the total document number of $R_q$, respectively. F-measure (Eq. [4]) combines recall and precision and focuses on the capability of detecting positives. Balanced accuracy (Eq. [5]) is used considering imbalanced true labels, which normalized true positive and true negative predictions by the number of positive and negative samples, respectively, and divided their sum by two.

$$Recall_q = \frac{\sum_{d \in R_q} rel_q(d)}{|D_q|} \tag{2}$$

$$Precision_q = \frac{\sum_{d \in R_q} rel_q(d)}{|R_q|} \tag{3}$$

$$F_q = \frac{2}{Recall_q^{-1} + Precision_q^{-1}} \tag{4}$$

$$Balanced\ Accuracy_q = \frac{\left(1 - \frac{|D_q| - \sum_{d \in R_q} rel_q(d)}{|D_q|}\right) + \left(1 - \frac{|R_q| - \sum_{d \in R_q} rel_q(d)}{A - |D_q|}\right)}{2} \tag{5}$$

# 4. EXPERIMENTS AND RESULTS

## 4.1 Experiments

Following the proposed pipeline, we evaluate the effectiveness of different NLP techniques in extracting semantics from GIS knowledge texts and uncover their semantic relevance. The latest version of GIS&T BoK (2021 3rd quarter)[1] at the time of writing is used. At Step 1, BoK topics are categorized into ten knowledge areas, including Foundational Concepts (FC), Knowledge Economy (KE), Computing Platforms (CP), Computer programming and Development (PD), Data Capture (DC), Data Management (DM), Analytics and Modeling (AM), Cartography and Visualization (CV), Domain Applications (DA), GIS&T and Society (GS). These topics follow a structure containing several segments. In our work, the most interesting segments of *Title, Keyword, Abstract*, *Main*, *Learning Objectives* (*LearnObj* for short),

---

[1] https://gistbok.ucgis.org/archives/2021-quarter-03

*Instructional Assessment Questions* (*IAQ* for short), are selected for analysis. *Abstract* is the author summary of a topic, usually containing one or two paragraphs. *Main* is the main body of a BoK topic, which contains detailed descriptions of this topic. *Learning Objectives* are listed in bullet points. *Instructional Assessment Questions* serve as a way for evaluating student learning outcomes.

A topic example can be found as "CP-14 - Web GIS" where CP denoted the knowledge area "Computing Platforms", 14 is the topic identifier, "WebGIS" is the topic name, it is about a technique "allows the sharing of GIS data, maps, and spatial processing across private and public computer networks". After deciding on segments of interest, we crawl and parse these textual segments from GIS&T BoK websites. At the end of a GIS&T BoK topic, there are some related topics listed by editors or writers. On average, each topic lists three related topics and they are used as the ground-truth labels to evaluate the effectiveness of the machine ranked results. Note that, the list of relevant topics is longer for some BoK entity and shorter for others, creating an imbalanced training dataset.

Six models, LSA, GloVe, RoBERTa, SPECTER, SciBERT, and Doc2vec are used to conduct semantic analysis and identify similar topics. The Gensim library[2] is employed for LSA and related text preprocessing. A key parameter of LSA model is the number of topics. Based on the GIS&T BoK hierarchy, the amount of knowledge area (10) and units (53) are both tested as the topic number, and setting topic numbers as 53 always yields better performance. For GloVe model, the common version with 840B tokens, 2.2M vocabularies, cased, and 300 dimensional vectors as output is used from Hugging Face[3]. The Sentence Transformers library[4] is used for RoBERTa and SPECTER models where all-roberta-large-v1, allenai-specter were used respectively. The required inputs of SPECTER model are the title and abstract of a paper, so the corresponding sections in the BoK topic is sent to the model. SciBERT was from the Allen Institute for AI[5] in the version of scibert_scivocab_uncased. For Doc2vec, the preprocessing is the same as what is applied to LSA. We pass all the GIS-T BoK text by the Gensim phrases module to form bi-grams and tri-grams to capture some phrases such as "remote sensing", "geographic information system", and so on. Stopwords such as "the", "a" are removed, and the rest are tokenized and passed to the Gensim Doc2vec[6] for training. To implement the KACERS summarizer, a cross-encoder of BERT with 2 layers pretrained on the MS MARCO dataset[7] is employed. Although other larger pretrained models are available for better performance but they run much slower than the 2-layer version, so here we use a simpler model to demonstrate the feasibility of KACERS.

Of the evaluation metrics, recall is the most important measure as the goal is to find more true positives between human proposed related topics and that identified by the machine. Balanced accuracy is the second most important metric because it reflects the overall capability of models especially considering imbalanced data. F-score and Precision are also calculated, but they are expected to decrease along with the increase of *k* (the number of candidate topics ranked by similarity scores), especially after all true labels are found. To answer the research question, model performance comparison and outputs analysis are reported in section 4.2.

---

[2] https://radimrehurek.com/gensim/models/tfidfmodel.html
[3] https://huggingface.co/sentence-transformers/average_word_embeddings_glove.6B.300d
[4] https://www.sbert.net
[5] https://huggingface.co/allenai/scibert_scivocab_uncased
[6] https://radimrehurek.com/gensim/models/doc2vec.html
[7] https://www.sbert.net/docs/pretrained_cross-encoders.html

## 4.2 Results

### 4.2.1 Model performance on original segments

We first evaluate how much semantics each original segment carries for performing semantic similarity comparison of two topics and which model provides the best recall (the highest consistency between the human and machine rankings). There are three categories of segments considering length and content. The first category is short segments including *Title* and *Keyword*. Although both segments provide phrases rather than complete sentences, they contain the most semantically expressive keywords about a topic. Intuitively, *Title* conveys very limited semantics because short titles are preferred to make the topic general and easy to understand in the BoK project.

Figure 3 lists the best-performed model taking different segments as input for the semantic analysis. As seen, RoBERTa works best in extracting most similar topics based on *Title* and *Abstract* (first and third column) because this model is pretrained on a lot of general corpuses which helps to understand the semantic meanings of short texts. The GloVe model using *Keyword* as input yields higher recall than RoBERTa taking *Title* as input, because GloVe takes advantage of aggregated word-word cooccurrence, which can represent the keyword-keyword cooccurrence in BoK entities very well. Compared to *Title* and *Keywords*, *Abstract* contains more semantics about a topic, hence the recall rate using *Abstract* for the semantic analysis is higher than using the other two segments. RoBERTa achieves the best performance than the other models when processing information in the *Abstract*, due to its novel strategy of attention-based context understanding. This model, however, does not work as well as Doc2Vec when processing *Main* (containing content in both *Abstract* and *Main* segments), because (1) the two segments together contain the most semantics of a topic; and (2) Doc2Vec is good at processing long texts to extract semantically meaningful features while the other models, such as RoBERTa, have a limit in the input length. Not surprisingly, models taking *LearnObj* and *IAQ* as input do not show as good performance (in terms of recall rate) as others, because the contents in *LearnObj* and *IAQ* are mostly about information used to support teaching and learning, so the semantics in them may not be directly related to a topic.

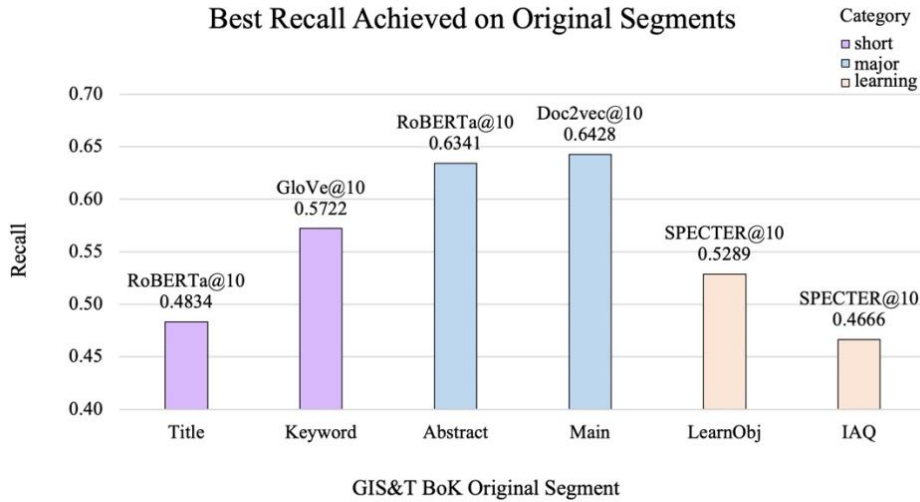

Figure 3. Comparison of model performance taking different BoK segments as input. This result is based on the top 10 (@10) most similar topics ranked by the best-performed models using some segment (e.g., *Title*) as input.

From the above analysis, we can tell that, the RoBERTa model which takes a much shorter text -- the Abstract as input yields almost as good performance as the Doc2Vec model taking nearly the entire document as input. Two issues limit the performance of the RoBERTa model here: (1) although the *Abstract* contains semantic description about a topic, the level of semantics is still confined to its length (e.g., 100 words); (2) RoBERTa is better at processing short text, and it limits the input token (i.e., a sequence of characters as a semantic unit) size to be 512. That means, input longer than 512 tokens will be truncated, resulting in a loss of data semantics. This motivates us to conduct further semantic analysis, especially semantic summarization of the original text to use a much shorter but near semantically-equivalent summary as the input data to fully utilize strengths of the RoBERTa model. Next section describes results based on our proposed method.

4.2.2 Model performance on semantic summary

Following the motivation stated above, the goal of this subsection is to generate a new topic summary containing more semantics than any of the individual segments in the original document to empower the semantic analysis by multiple models. The idea of semantic summary is to conduct a semantic summarization using our proposed KACERS summarizer on both *Abstract* and *Main*, and use the combined summary as input for topic relevance analysis. The KACERS summarizer can select the top n (n=1, 2, 3...) most similar sentences in a text segment given any keyword, which allows us to generate multiple semantic summaries when n is set to different values. Table 1 shows the recall rate achieved using the RoBERTa model by taking different semantic summary as input. The row and column labels show the parameters (n=1,2,3) used in KACERS to generate the semantic summary from *Main* and *Abstract* respectively. For instance, the cell in the third row and third column has a value of 0.6578; it means that by using RoBERTa and taking the combined summarized *Main* (n=1 in KACERS) and original *Abstract* as input, the best recall is achieved based on top 10 most relevant topic and the score is 0.6578. Column-wise, the improvement in recall (% value in parentheses) compared with the first observed value (using *Abstract* alone) is presented. The improvement is obvious with injections of the summary from *Main*, as it provides additional, useful semantics. The improvement is observed even directly attaching the summarized *Main* produced by KACERS (at n=1, 2, or 3) to the original *Abstract* (values in the third column). However, this column does not act as the overall best summarizer, indicating that a balance between deliveries of *Abstract* and *Main* is important. The summarized *Abstract* obtained by KACERS at n=2 combined with the summarized *Main* obtained by KACERS at n=3 yields this the highest recall score of 0.6927. And this result is also better than that using any original segment as input (Figure 3). We name this best summary combination as *Semantic Summary* for later reference.

Table 1. Recall@10 performance comparison of different combinations of KACERS (keyword-aware cross-encoder-ranking summarizer) applications on *Abstract* segment and *Main* segment.

|  | Main Only | + Original Abstract | + Summarized Abstract (KACERS-1) | + Summarized Abstract (KACERS-2) | + Summarized Abstract (KACERS-3) |
| --- | --- | --- | --- | --- | --- |
| Abstract Only | / | 0.6341 | 0.4466 | 0.5041 | 0.5526 |
| Summarized Main (KACERS-1) | 0.5948 | 0.6578 ( ↑ 3.74%) | 0.5909 ( ↑ 32.31%) | 0.6192 ( ↑ 22.84%) | 0.6105 ( ↑ 10.48%) |
| Summarized Main (KACERS-2) | 0.6577 ( ↑ 10.58%) | 0.6491 ( ↑ 2.36%) | 0.6341 ( ↑ 41.99%) | 0.6527 ( ↑ 29.48%) | 0.662 ( ↑ 19.81%) |
| Summarized Main (KACERS-3) | 0.6295 ( ↑ 5.84%) | 0.6776 ( ↑ 6.86%) | 0.6386 ( ↑ 42.98%) | **0.6927** ( ↑ 37.41%) | 0.6542 ( ↑ 18.39%) |

*KACERS-n: top n sentences in the input segment of each input keyword, then truncated as the summary.
*The increase percentage is based on the first observed recall in the same column.

Also, *Semantic Summary* generated by KACERS is compared with combinations of related segments - *Title*, *Abstract*, *Keyword*, and *Main* - without KACERS (Figure 4). The first column shows the best *recall@10* value achieved by GloVe by taking the combination of short segments (*Title* + *Keyword*) as input. As seen, the recall rate (0.5859) is better than using either original segment as input (respective results are shown in Figure 3), and this is again consistent with our intuition that those two segments are valuable and compensate with each other. GloVe using the combined *Title* and *Keyword* as input outperforms other models because data in this combination contains words and phrases where word-word cooccurrence plays an important role in semantic similarity. The second column lists the recall rate by taking longer segments (*Title* + *Keyword* + *Abstract* + *Main*) as input. RoBERTa model achieves the best *recall@10* on this combination. It again surpasses results using any individual segment as input (Figure 3). RoBERTa achieves the overall best recall rate using the *Semantic Summary* as input because (1) as summarized text carries more semantic information, and (2) attention-based context understanding mechanism adopted in RoBERTa is powerful in conducting semantic analysis of short texts.

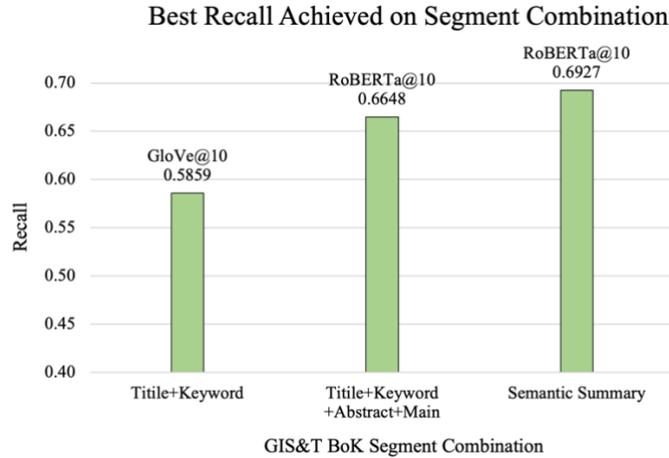

Figure 4. Best models on different segment combinations considering recall. @10 means on the top 10 most similar topics ranked by the best-performed models using some segment combination.

Figure 5 further illustrates the performance comparison of different deep learning model using the Semantic Summary as the input. X axis shows the increase in $k$ when it is used as the parameter to select the top ranked similar topics by the machine to conduct the evaluation. Y axis shows the performance score using different metrics. In general, the best model achieves the highest recall and balanced accuracy at the same time. The best recall 0.6927, is achieved by model RoBERTa when $k$=10. It also has the highest balanced accuracy when $k$=10, 0.8161. In other words, the *Semantic Summary* serves as a great semantic representation of a BoK topic. The results can also tell that when $k$ increases from 1 to 10, both recall and the balanced accuracy increase. This trend shows that NLP models get better performance if we allow them to suggest more similar topics. In the "Recall" subplot, RoBERTa outperforms the other models on almost all choices of $k$. For models pre-trained in scientific literature, SPECTER works better than SciBERT. Although SciBERT claims to achieve high scores on some NLP tasks, it was originally trained on biomedical domain papers. This pretraining bias is exposed in this GIS domain task. LSA generally performs better than GloVe. The text length of *Semantic Summary* is short for Doc2vec to achieve a good performance.

Models excel in precision when fewer top $k$ sentences suggested by the model is considered, and then trend down as $k$ increases. The reason is that the related topic list is on average short as mentioned in section 4.1, containing only a few topics (e.g., 2-3). Models achieve over 40% precision when $k$=1 reflecting that there is a high chance that they can correctly find at least one same topic as provided by the topic authors. The more relevant candidates identified by a machine is considered, the more topics will be found. Thus, the precision score drops quickly. This short relevance characteristic also explains the abrupt summit of F-score near $k$=2 for most models. After $k$=2, the declination of precision is higher than the increase of recall, so we can observe a drop in the F-score. Based on this, precision and F-score are not very meaningful metrics for this GIS&T BoK similarity task when $k$ is high, but recall and balanced accuracy are still valuable.

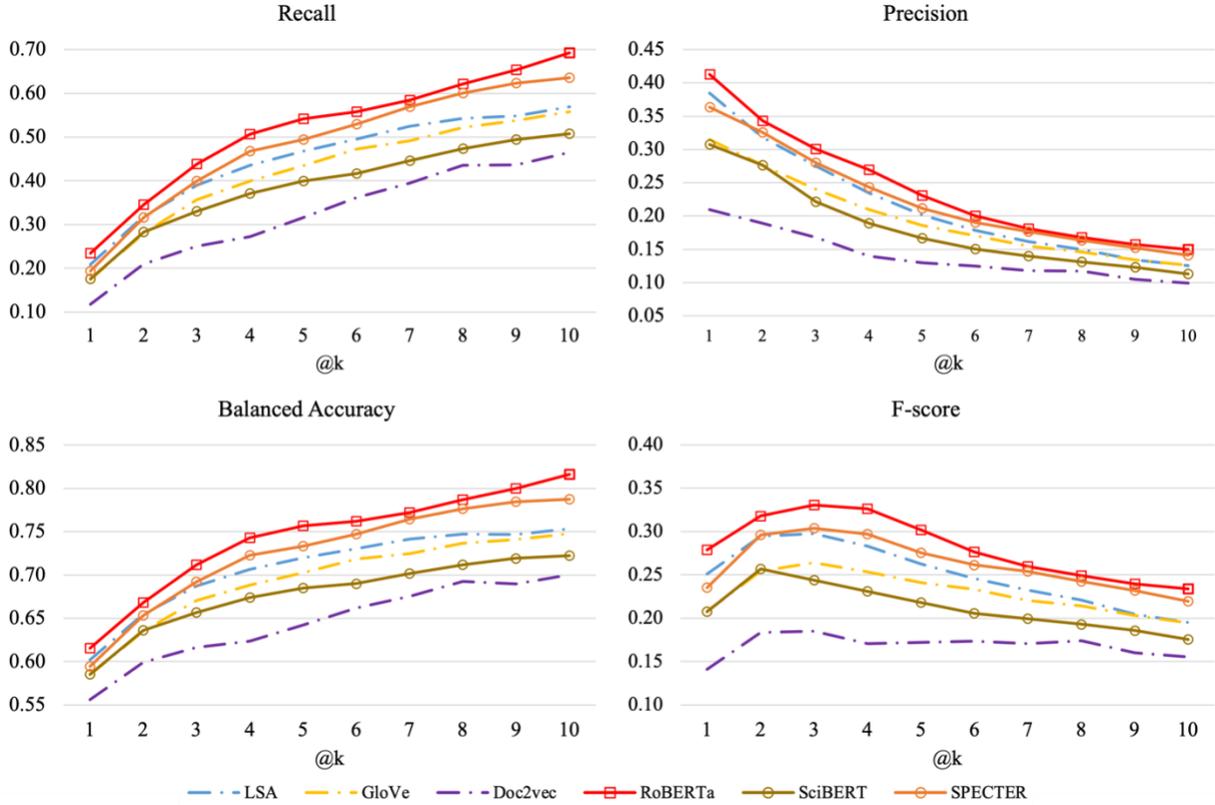

Figure 5. Model performance comparison taking *Semantic Summary* as the input for similarity analysis.

4.2.3 A running example for the machine-ranked and human-ranked similar topics

We use NLP models to find the relevance between GIS&T BoK topics and provide top *k* related topics. As compared above, *Semantic Summary* produced by KACERS serves as the best semantic representation, and RoBERTa is the best model on that achieves the highest *recall@10* for similarity ranking. Here we provide an example to show the differences in machine-ranked results and author-identified relevant topics. According to the machine ranking, the ten most similar topics of "GS-14 - GIS and Critical Ethics" (i.e., $R_{q=GS\text{-}14}$) are listed in Table 2. This (seed) topic discusses critical ethics and GIS, exploring how spatial data can mask, conceal or disallow knowledge, and suggests using concepts from educational pedagogy to enrich GIS experience, pointing to the practical, empirical and political nature of ethical critique. This is an important research area that may engage with interdisciplinary debates.

Here, some related topics ignored by humans are found by our model, such as the top-ranked topic "GS-11 - Professional and Practical Ethics of GIS&T". There is a difference between ethics in GS-14 and ethics in GS-11 because the latter is more on the operational level, but they are closely related. As for consistency, "GS-13 - Epistemological critiques" and "GS-15 - Feminist Critiques of GIS" fulfill the related topic list provided by humans. In this case, *recall@4* reaches 100%.

Referring to BoK's hierarchical structure[8], we can further analyze similar topics within the hierarchy. As introduced in Section 1, GIS&T BoK has an inherent hierarchy, which includes knowledge areas, units, then topics. In Table 2, half of the top 10 similar topics and the top four most similar topics are all from the GIS&T and Society (GS) knowledge area as the seed topic, two of which are identified by the authors. The algorithm also finds similar topics from a number of other knowledge areas, such as "DA-11 - GIS&T and the Digital Humanities" from Domain Application (DA) knowledge area. This demonstrate that NLP has the potential assisting people to uncover more possibilities of connectivity than we used to think about, where the latent similarity exists across units and knowledge areas.

Table 2. An example of top $k$ related topic candidates ($R_q$) ranked by machine for a seed topic. $q$ = "GS-14 - GIS and Critical Ethics", $k$ = 10.

| Seed Topic | Similar Topic Ranked by the NLP Model | Similarity Rank | Identified by Human |
|---|---|---|---|
| GS-14 - GIS and Critical Ethics | GS-11 - Professional and Practical Ethics of GIS&T | 1 | No |
| GS-14 - GIS and Critical Ethics | GS-13 - Epistemological critiques | 2 | **Yes** |
| GS-14 - GIS and Critical Ethics | GS-12 - Ethics for Certified Geospatial Professionals | 3 | No |
| GS-14 - GIS and Critical Ethics | GS-15 - Feminist Critiques of GIS | 4 | **Yes** |
| GS-14 - GIS and Critical Ethics | DA-11 - GIS&T and the Digital Humanities | 5 | No |
| GS-14 - GIS and Critical Ethics | CV-26 - Cartography and Power | 6 | No |
| GS-14 - GIS and Critical Ethics | FC-03 - Philosophical Perspectives | 7 | No |
| GS-14 - GIS and Critical Ethics | GS-04 - Location Privacy | 8 | No |
| GS-14 - GIS and Critical Ethics | FC-24 - Conceptual Models of Error and Uncertainty | 9 | No |
| GS-14 - GIS and Critical Ethics | FC-35 - Openness | 10 | No |

*Topic naming: knowledge area name abbreviation followed by a unique topic ID within this knowledge area.

### 4.2.4 Analysis on intra-KA similarity and inter-KA similarity of machine ranked similar topics

The following analysis provides a comprehensive view of intra-KA similarity and inter-KA similarity of GIS&T BoK topics. Although the current BoK design has grouped the topics into distinct categories, related topics can still be found across different knowledge areas, as demonstrated by both human labels and NLP model-generated candidates. This is reasonable considering all the topics are in the GIS domain, which is interdisciplinary in nature. Figure 6 presents two directed chord plots and a bar chart that visualize semantic similarity of topics within and across knowledge areas. Knowledge areas are arranged radially around a circle and relationships in between are drawn as chords. A chord and the seed topic's knowledge area share the same color, for instance, a blue chord between "Foundational Concepts" in blue and "Domain Applications" in green indicates that a seed topic from "Foundational Concepts" has a relevant candidate belonging to "Domain Applications". If a knowledge area has chords curving to itself, then there exists intra-similarity of tropics within knowledge area. Comparing similarity@1 (Figure 6a) and similarity@10 (Figure 6b), we notice that the overall chord distribution pattern looks alike, while the latter has more dense chords. Despite the observed inter-similarity and intra-similarity, the interaction of similarity may vary among different knowledge areas.

---

[8] https://ucgis.memberclicks.net/assets/docs/BoK/Topic_Overview_9_30_2022.pdf

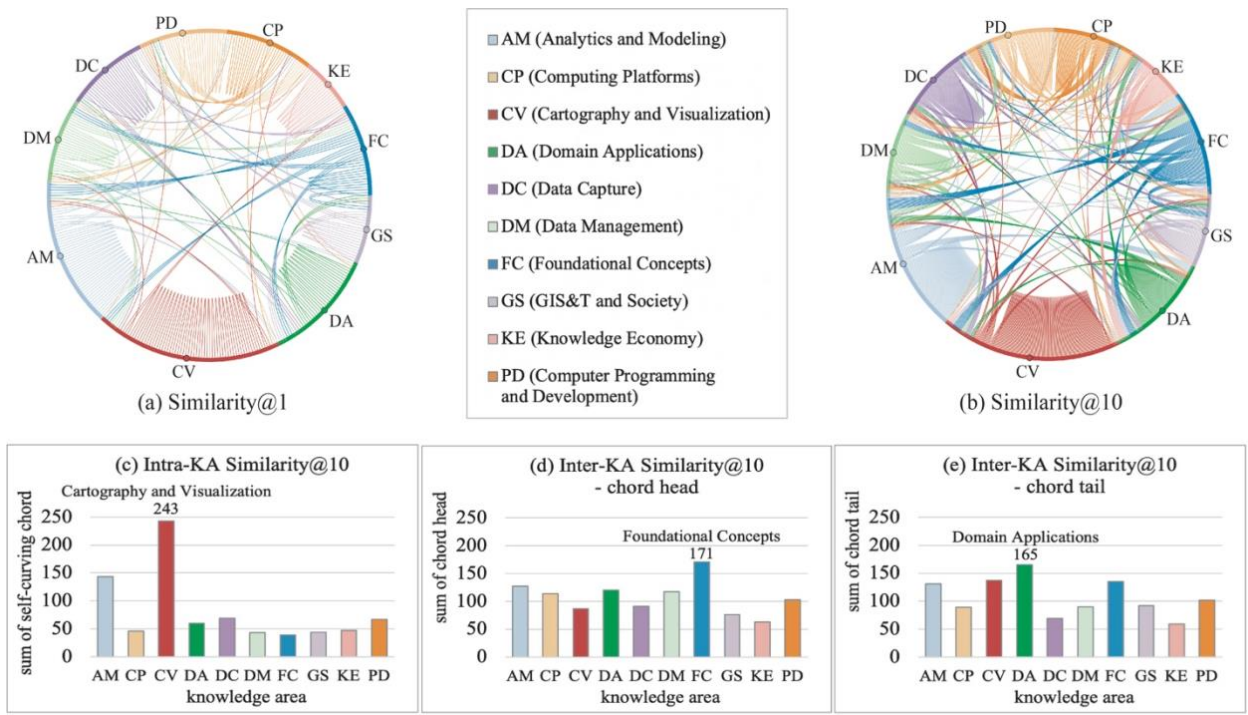

Figure 6. Intra-similarity and inter-similarity of knowledge areas among the GIS&T BoK topics.
(a) chord diagram of similarity@1; (b) chord diagram of similarity@10;
(c) intra-KA similarity@10, sum of self-curving chords per KA;
(d) inter-KA similarity@10, sum of chord heads per KA;
(e) inter-KA similarity@10, sum of chord tails per KA.

Figure 6c presents statistics of similarity@10. The knowledge area "Cartography and Visualization" (red bar) has the highest intra-KA similarity, indicating that topics within area are highly interconnected. Additionally, to determine the highest inter-similarity between knowledge areas, we compared the sums of the chord heads and chord tails, respectively. A head is a seed topic, and a tail is a similar candidate of the seed topic. An interesting observation is that GIS&T BoK topics tend to start from "Foundational Concepts" and end at "Domain Applications". This suggests that our model has effectively captured the semantics and human knowledge within the this BoK, as it reflects the general understanding that a domain typically expands from concepts to applications.

## 5. DISCUSSION

Our approach has several potential applications including assisting BoK authors and editors to quickly create and update of the "Related Topics" section of each entry. We recommend that when the BoK project reaches a stable stage, each topic author will be contacted with several suggested topics of high relevance identified by our proposed machine learning model. A list of top 10 related topics would be a good starting point. Our finding (Figure 4) suggests that there should be a near 70% chance that the top 10 topic list contains results highly consistent with the human labels. The authors can rank according to their knowledge and expertise which are relevant and which are not. These answers will be very valuable for further analysis when both relevance and irrelevance topics are logged.

There are some limitations of the proposed work. The good performance of KACERS relies on high-quality keywords. However, in order to provide high-quality keywords, a topic author should have solid domain knowledge and minimize human bias. The quantity, scope, and expandability of keywords will affect quality of keywords. When the BoK expands and includes new topics, the chosen set may be constrained if the keywords to be chosen are predefined. Even though the keyword list is extendable, choosing whether to include a new keyword that is on the margins of GIS might be challenging. This ambiguity is obvious when authors address topics in Domain Applications that are of specific interest. It could be amplified when authors come from a marginalized group of scholars. This uncertainty will have an effect on both the summary process and the semantic similarity.

Another point worth mentioning is that, although consistent with domain experts' intuition, the learning objective segment and instructional assessment question segments are not suitable for the semantic similarity task, instead their semantics can be used for other purposes. For example, sentences in *Learning Objectives* usually start with action verbs such as "explain", "distinguish". Sentences in *Instructional Assessment Questions* usually start with question verbs such as "how", "what", "why". Those words imply interconnections between notions mentioned in the text. If we can recognize those notions and link them to topics, then it is possible to present more relationships between topics and further support the different knowledge exploration needs.

# 6. CONCLUSION

This work introduces several state-of-the-art models to explore the semantic relations between knowledge topics to advance relevance discovery in body of knowledge. The GIS&T Body of Knowledge project is used as a use case. Our paper makes three contributions. First, a BoK topic has many segments and the level of semantics expressed in each of these segments is quite different. To better understand how well they can deliver the core message of the topic, the semantic analysis of different topic segments is conducted. Second, in order to create a better semantic representation of a topic, a novel summarizer called KACERS is created. KACERS successfully completes the challenge of comparing the semantics of long texts and produces a semantic summary that is both an accurate semantic representation of the original text and substantially shorter in length. This way, it allows NLP models which have limits in input data size to better demonstrate their power. A reduced input data amount also makes the machine learning process more efficient. The generated semantic summary is also very easy to comprehend and evaluate. Third, we implement various models to extract semantics from different segments of BoK topics. Effectiveness of traditional NLP models, generally pre-trained models and academic domain pre-trained models are compared and evaluated. We discuss model performance for each segment and give insights on pairwise relevance based on the best model.

Our experiments demonstrated the power of pre-trained natural language processing models to understand semantics of BoK. The proposed framework can also assist iterations of other geospatial BoKs such as the Encyclopedia of Mathematical Geosciences, or GIScience scholars who want to summarize or compare semantics of acknowledged geospatial knowledge. There has been a pursuit of achieving higher performance based on custom models in the past years. However, utilizing pre-trained models shouldn't be ignored (Qiu et al., 2020). A general purpose of a body of knowledge is to serve people with different knowledge backgrounds, which means their understanding of text is more general than a domain expert, which influences their relevance judgement. Moreover, it is easier to scale up the relevance retrieval capability with more incoming topics if the model is out-of-the-box and ready to use with a good performance. We vision the ultimate goal of various bodies of knowledge shouldn't be confined within a specific domain. If we want to build a comprehensive knowledge graph consisting of BoKs from different domains such as a comprehensive body of geography, urban planning, and economics, a general pipeline built upon pre-trained model can serve as an efficient solution to discover relevance between topics, ensuring its transferrability and reproducibility (Goodchild & Li, 2021). And here, we demonstrate the potential of deep learning models, such as RoBERTa, in achieving this goal.

# [References]